# Methods of ranking for aggregated fuzzy numbers from interval-valued data

Justin Kane Gunn
School of Computing and Information Systems
The University of Melbourne
Melbourne, Australia
justinkanegunn@gmail.com

Hadi Akbarzadeh Khorshidi
School of Computing and Information Systems
The University of Melbourne
Melbourne, Australia
hadi.khorshidi@unimelb.edu.au

Uwe Aickelin
School of Computing and Information Systems
The University of Melbourne
Melbourne, Australia
uwe.aickelin@unimelb.edu.au

*Abstract*—This paper primarily presents two methods of ranking aggregated fuzzy numbers from intervals using the Interval Agreement Approach (IAA). The two proposed ranking methods within this study contain the combination and application of previously proposed similarity measures, along with attributes novel to that of aggregated fuzzy numbers from interval-valued data. The shortcomings of previous measures, along with the improvements of the proposed methods, are illustrated using both a synthetic and real-world application. The real-world application regards the Technique for Order of Preference by Similarity to Ideal Solution (TOPSIS) algorithm, modified to include both the previous and newly proposed methods.

I. INTRODUCTION

It is common practice for experts to summarize their opinions into a single numerical score, for example, "an 8 out of 10" as seen throughout media reviewing along with academic and job candidate selection. The alternative of summarizing opinions as intervals, for example "a 6 to 8 out of 10" in contrast to utilizing singular real numbers, allows for the representation of uncertainty, of which is the inspiration for its growing research interest. As with traditional scoring, interval scores can be aggregated into sets as to provide a better reflective summary, either as intra-person or inter-person variability, with intra-person referring to an expert's opinion changing over time and inter-person referring to a group of different expert opinions [1, 2, 3].

Recent literature explores the construction of fuzzy numbers from interval-valued datasets. As fuzzy numbers are data types developed with the intention of capturing uncertainty [4] and have a rich research history, utilizing them as a means of representation for interval-valued datasets allows for further convenient and potential applications. Liu and Mendel [5] first proposed a widely used method of converting a set of intervals to fuzzy numbers, known as the Interval Approach (IA). Further improvements were proposed by Coupland et al. [6], which was referred to as the Enhanced Interval Approach (EIA), and the method proposed by Miller et al. [3] which enabled both the use of intra-expert and inter-expert opinions and the removal of preprocessing the input data. Wagner et al. [1] building upon previous literature introduces the most recent major method for converting interval sets to fuzzy numbers, known as the Interval Agreement Approach (IAA). The IAA method offers a refined approach that includes the benefits of the method proposed in [3], and furthermore is shown to intentionally not make any assumptions about the input interval-valued dataset. Thus, IAA produces a wide and unrestricted variety of fuzzy numbers in contrast to the previous interval to fuzzy number methods; which in theory fully represents and maintains the uncertainty that the intervals were initially intended to provide. Whilst expert opinions are the primary example data utilized throughout this research area, IAA can produce fuzzy numbers regardless of the context of the interval-valued input data [2]. We refer to aggregate fuzzy numbers from interval-valued data using the IAA method as IAA Fuzzy Numbers.

Due to IAA Fuzzy Numbers being a relatively recent data type and along with their flexibility, research is still within the process of having the IAA Fuzzy Number primary mathematical operations universally defined and in practice. Whilst IAA Fuzzy Numbers provide a novel means of capturing uncertainty, without the standard operations defined, they remain difficult to apply and analyze within modern systems. The objective of this study is to propose further operations of IAA Fuzzy Numbers, with the motivation that IAA Fuzzy Numbers may be potentially applied to algorithms. Common operations relevant to the application of many data types are methods of similarity and ranking, as they remain a requirement for many contemporary algorithms, and are involved in many areas including that of decision-making, pattern-matching and classification. Similarity measures are a means quantifying the similarity between two instances of a specific data type, whilst ranking methods evaluate the rank (i.e. order) of two or more instances of a specific data type. Wagner et al. [7] propose a similarity measure for Type-1 Fuzzy Numbers which utilizes the Jaccard similarity coefficient and is applicable to IAA Fuzzy Numbers [7, 8]. In contrast to [7], Gunn et al. [9] propose a similarity measure for IAA Fuzzy Numbers that utilizes a collection of attributes as features, along with each weight of said feature calculated by Principal Component Analysis (PCA); this alternative technique was inspired by Khorshidi and Nikfalazar's similarity measure for Generalized Fuzzy Numbers in [10]. Note that the proposals of [9] are in their current state only applicable to Type-1 fuzzy numbers. Various types of fuzzy numbers have seen proposals for both similarity measures and ranking methods [10, 11, 12, 13], which has provided opportunities for their further practical application. However, previous to this study, there has not been a method of ranking for IAA Fuzzy Numbers formally proposed within the surrounding literature. Whilst it is common that similarity measures may sometimes be also applied as methods of ranking, the similarity measures of previous IAA literature have shortcomings in this regard and require modification. Building upon the findings in [7, 9], we propose a new similarity measure and two ranking methods for IAA fuzzy numbers and provide both an update to the previous illustrative example in [9], as well as explore the results from the application in a real-world case study [14, 15] as a means to demonstrate their potential use.

This paper is organized as follows. Section II provides a brief background on the topics addressed within this paper. Section III demonstrates the shortcomings of the previous IAA Fuzzy Number similarity measures when applied to the area of ranking. Section IV outlines this study's proposed methods of similarity and ranking for IAA Fuzzy Numbers, and demonstrates the prior shortcomings now addressed using said proposed methods. Section V explores the potential applications of the proposed methods using real-world data. Section VI concludes this paper with a discussion on the findings of this study and potential future work.

## II. BACKGROUND

In this section, we provide a *very* brief introduction into the IAA method for Type-1 fuzzy numbers [1, 2], and include the outlines of the similarity measures proposed in [7, 9].

### A. IAA Fuzzy Numbers

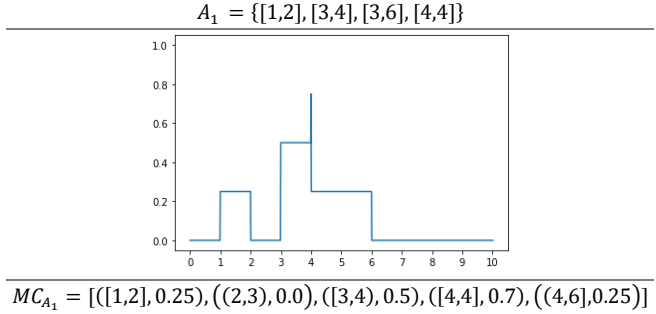

Fig. 1. Output of IAA given the set of intervals $A_1$, along with its reflective $MC$ list [9]

A crisp interval is defined as $\bar{A} = [l_{\bar{A}}, r_{\bar{A}}]$, where $l$ is the 'left' start value and $r$ is 'right' end value, and thus a set of these intervals is defined as $A = \{\bar{A}_1, ..., \bar{A}_n\}$. Given set $A$ as input, IAA [1, 2] is formulated as (1). Where $\mu_A$ is the membership function, and $y_i = i/n$ is the degree of membership. Note that the use of the solidus " / " in (1) is not referring to a division, but instead to the assignment of a given degree of membership to a set of values. The variable $y_i$ only equals 1 at values in which all intervals of set $A$ overlap.

$$\mu_A = \sum_{i=1}^{n} y_i \Big/ \Big(\bigcup_{j_1=1}^{n-i+1} \bigcup_{j_2=j_1+1}^{n-i+2} \cdots \bigcup_{j_i=j_{i-1}+1}^{n}(\bar{A}_{j_1} \cap ... \cap \bar{A}_{j_i})\Big) \quad (1)$$

In the context of Type-1 Fuzzy Numbers, the original definition of IAA (1) can further be simplified into (2, 3) [1, 2]. This variant of IAA defines the degree of membership of a variable real number $x$ for a given set of crisp intervals $A$, as the frequency (or count) of which $x$ appears within each interval divided by $n$ (i.e. the cardinality of set $A$). Whilst $x$ can be any real number, literature commonly refers to $x$ in the context of $x \in X_{\bar{A}}$, in which $X_{\bar{A}}$ is the ordered set of all $l_{\bar{A}}$, and $r_{\bar{A}}$ [2].

$$\mu_A(x) = \Big(\sum_{i=1}^{n} \mu_{\bar{A}_i}(x)\Big)/n \quad (2)$$

$$\mu_{\bar{A}_i}(x) = \begin{cases} 1 & l_{\bar{A}_i} \leq x \leq r_{\bar{A}_i}, \\ 0 & else. \end{cases} \quad (3)$$

Refer to Fig. 1 for an example output of IAA given an elementary set of intervals. Note the list $MC_A$ is a method of representing an IAA Fuzzy Number proposed in [9] as a means of convenient data gathering, it is a list of tuples that record the boundaries of each region over the membership function $\mu_A$ curve, an abstract definition is given by (4, 5). Note a region $R_i$ is a line when $R_{i_l} = R_{i_r}$.

$$R_i = ([R_{i_l}, R_{i_r}], R_{i_h}) \quad (4)$$

$$MC_A = [R_1, R_2, ..., R_{n-1}, R_n] \quad (5)$$

### B. Similarity Measure: Jaccard similarity coefficient

Wagner et al. [7] propose a similarity measure applicable to Type-1 IAA Fuzzy Numbers, which applies the Jaccard similarity coefficient, also known as the 'intersection over union' between two sets. The Jaccard similarity coefficient is a proportional value $\in [0, 1]$ that represents how similar set A is to set B, with 1 indicating that they are identical and 0 indicating they are disjoint [7, 8]. In the case of Type-1 Fuzzy Numbers, the Jaccard similarity coefficient as described in [7, 8] is given as (6). In the case of two input IAA Fuzzy Numbers, $\tilde{A}$ and $\tilde{B}$, the default definition for all $x_i$ is $x \in (X_{\tilde{A}} \cup X_{\tilde{B}})$.

$$S_J^{CS}(\tilde{A}, \tilde{B}) = \frac{\tilde{A} \cap \tilde{B}}{\tilde{A} \cup \tilde{B}} = \frac{\sum min(\mu_{\tilde{A}}(x_i), \mu_{\tilde{B}}(x_i))}{\sum max(\mu_{\tilde{A}}(x_i), \mu_{\tilde{B}}(x_i))} \quad (6)$$

### C. Similarity Measure: Attribute Comparison

Gunn et al. [9] propose a similarity measure that combines six comparative attribute features along with their respective weights into a single measure; the weights were derived from the application of Principal Component Analysis (PCA) [16] on a large random data set of IAA Fuzzy Numbers. This similarity measure is outlined by (7) and Table. I. Refer to [9] for a full description of each attribute. Like the previous similarity measure, the output is a proportional value $\in [0, 1]$ representing the similarity between two IAA Fuzzy Numbers $\tilde{A}$ and $\tilde{B}$.

$$S_{AC}(\tilde{A}, \tilde{B}) = \Big(1 - \sum_{i=1}^{6} w_i^2 f_i\Big) \quad (7)$$

TABLE I. FEATURES AND WEIGHTS FOR SIMILARITY MEASURE [9]

| Distance / Difference Measure | Feature vector – $f$ | Weight vector - $w$ |
|---|---|---|
| Quartile | $\dfrac{\sum_{i=1}^{5}\|a_i - b_i\|}{5(range)}$ | 0.320726 |
| Centroid | $\dfrac{\sqrt{(Cx(\tilde{A})-Cx(\tilde{B}))^2 + (Cy(\tilde{A}) - Cy(\tilde{B}))^2}}{\sqrt{(range)^2 + 0.5^2}}$ | -0.509757 |
| Area | $\dfrac{\|A(\tilde{A}) - A(\tilde{B})\|}{\max(A(\tilde{A}), A(\tilde{B}))}$ | 0.100985 |
| Height | $\|w_{\tilde{A}} - w_{\tilde{B}}\|$ | -0.461649 |
| Perimeter | $\dfrac{\|P(\tilde{A}) - P(\tilde{B})\|}{\max(P(\tilde{A}), P(\tilde{B}))}$ | 0.444451 |
| Agreement Ratio | $\|AR(\tilde{A}) - AR(\tilde{B})\|$ | -0.465218 |

## III. SHORTCOMINGS

In this section, we describe the limitations of the previously outlined similarity measures (6, 7), especially concerning their potential applications to ranking IAA Fuzzy Numbers. Furthermore, we illustrate these limitations using an example given in [9].

### A. Similarity to Ranking

When attempting to transform a similarity measures to a method of ranking, an ideal best or ideal worst score must be included as a means of evaluating an increase in direction to a desirable state. Given some predetermined context involving sets of intervals, such as a case study, it is straightforward to evaluate the ideal best (i.e. highest possible potential result). For example, if 5 critics were ranking films out of 1 to 10, then the absolute best set of interval score responses a film could receive would be 5 intervals only containing 10 for both left and right points, that is $A_{best} = \{[10,10], [10,10], [10,10], [10,10], [10,10]\}$. In abstract terms, the ideal best is simply a set of intervals in which the participants (e.g. experts) have all provided the highest possible interval score, with the ideal worst being instead the lowest possible score. After computing the ideal IAA Fuzzy Numbers for a given scenario, each of the recorded IAA Fuzzy Numbers can simply be ranked by greatest similarity to the ideal best, or conversely and perhaps less intuitively, lowest similarity to the ideal worst. Ranking towards an ideal best is a general means of transforming a similarity measure to a ranking method. However, given that there exists an ideal worst that can be evaluated in the same manner for the majority of case studies, it would logically follow to include it within the ranking computation as it would achieve more effective results. To include both ideals within the evaluation, we simply divide the similarity between the IAA Fuzzy Number and the best ideal, with the sum of the similarity to both ideals. Refer to (12) for an equation that applies this concept.

Both similarity measures (6, 7) are effective at their primary tasks, (6) evaluates the magnitude in which two IAA Fuzzy Numbers overlap, whilst (7) is a general comparison of their overall structure, of which includes various attributes such as shape, position, and their maximum values. When ranking however, both similarity measures do have shortcomings. The Jaccard similarity measure (6) will return a value greater than zero only when the two IAA Fuzzy Numbers overlap at some point across the domain $X$; thus any two IAA Fuzzy Numbers that do not overlap in any way with an ideal will be considered equal in rank even when one of them is clearly skewed more desirably than the other. The attribute similarity measure (7) will return a value above 0 in virtually all cases. Equation (7) has a seemingly opposite issue to (6), in that it is highly sensitive to shape (similarity in patterns) and thus over ranks IAA Fuzzy Numbers that share said shape with an ideal even when they are distant from it on the x-axis (meaning they do not overlap at all).

### B. Illustrated Ranking Improvements

The following is a synthetic example that was produced and explored by [9] to illustrate the use of its attribute similarity measure. The primary aspects of this illustrative example will be explained in brief, refer to [9] for a further description of why this dataset was utilized.

The following dataset, Table. II, regards ten films that have been reviewed by five critics.

TABLE II. SYNTHETIC FILM REVIEW DATASET [9]

|  | Critic 1 | Critic 2 | Critic 3 | Critic 4 | Critic 5 |
|---|---|---|---|---|---|
| **Film A** | [1, 1] | [1, 1] | [1, 1] | [1, 1] | [1, 1] |
| **Film B** | [5, 6] | [6, 7] | [10, 10] | [3, 4] | [5, 5] |
| **Film C** | [2, 3] | [1, 3] | [4, 7] | [1, 3] | [4, 5] |
| **Film D** | [6, 6] | [6, 10] | [8, 10] | [5, 9] | [2, 3] |
| **Film E** | [1, 4] | [2, 3] | [7, 8] | [3, 3] | [2, 4.4] |
| **Film F** | [7, 7] | [8, 9.2] | [9, 10] | [8, 9] | [9, 10] |
| **Film G** | [8, 9] | [9, 10] | [9.5, 9.5] | [9, 10] | [10, 10] |
| **Film H** | [1.5, 6.5] | [3, 10] | [1, 10] | [2, 9.3] | [8, 8.8] |
| **Film I** | [8, 8] | [8, 8] | [8, 8] | [8, 8] | [8, 8] |
| **Film J** | [10, 10] | [10, 10] | [10, 10] | [10, 10] | [10, 10] |

Hypothetically, the critics were asked to give their opinion as an interval, of which crisp (i.e. when $l_{\bar{A}} = r_{\bar{A}}$) and decimals values were permitted, with scores being from 1 to 10 [9]. Regarding IAA Fuzzy Numbers, each film (i.e. row) within the dataset has a corresponding set of interval scores, these sets interval-valued data are to be used as input into the IAA method, thus producing a unique IAA Fuzzy Number for each film listed. Note that Film A is intentionally the ideal worst set of interval scores, whilst Film J is the ideal best. The output of IAA for this dataset is given as Fig. 2.

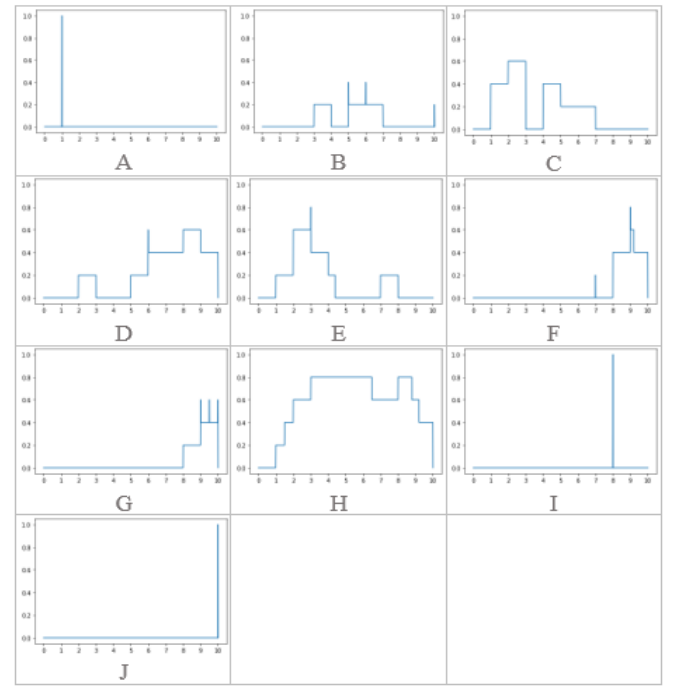

Fig. 2. Output of Film IAA Fuzzy Numbers

Review scores are commonly used to rank films, which aids individuals in deciding whether they are interested in a particular film; falling under the area of decision-making. The traditional method of ranking aggregated film scores would simply be by their respective average score. However, given that the dataset has been recorded in intervals as opposed to the norm of real numbers, when receiving such data a likely intuitive approach would be to preprocess it using the mean of each interval in contrast to using IAA Fuzzy Numbers. Computing the rank of each film using this traditional method allows for a means of contrasting and thus further evaluation of methods that utilize IAA Fuzzy Numbers.

The traditional averaging method using the mean of intervals as standard scores, is given in Table. III. As expected, the traditional approach evaluates Film A as the lowest ranking film, and Film J as the highest. Taking the averages of the preprocessed data allows for a rudimentary means of comparison. Whilst deviation may be evidence for potential new inferences utilizing the uncertainty of the data, large differences in ranking from the preprocessed averages would imply issues likely exist, especially when Film A and Film J's boundary rankings are not met.

TABLE III.  INTERVAL TO MEAN - FILM REVIEW DATASET

|        | Critic 1 | Critic 2 | Critic 3 | Critic 4 | Critic 5 | Average |
|--------|----------|----------|----------|----------|----------|---------|
| Film A | 1        | 1        | 1        | 1        | 1        | 1       |
| Film B | 5.5      | 6.5      | 10       | 3.5      | 5        | 6.1     |
| Film C | 2.5      | 2        | 5.5      | 2        | 4.5      | 3.3     |
| Film D | 6        | 8        | 9        | 7        | 2.5      | 6.5     |
| Film E | 2.5      | 2.5      | 7.5      | 3        | 3.2      | 3.74    |
| Film F | 7        | 8.6      | 9.5      | 8.5      | 9.5      | 8.62    |
| Film G | 8.5      | 9.5      | 9.5      | 9.5      | 10       | 9.4     |
| Film H | 4        | 6.5      | 5.5      | 5.65     | 8.4      | 6.01    |
| Film I | 8        | 8        | 8        | 8        | 8        | 8       |
| Film J | 10       | 10       | 10       | 10       | 10       | 10      |

With the traditional aggregated average scores as a reference, we now compute the similarity of each film to both ideals using the previously proposed similarity measures, given as Table. IV.

TABLE IV.  SIMILARITY TO RANKING METHOD OUTPUT (6, 7)

| Measure To Ideal | $S_J^{CS}(\tilde{F}, \tilde{I})$ -Eq. (6) | | | $S_{AC}(\tilde{F}, \tilde{I})$ -Eq. (7) | | |
|---|---|---|---|---|---|---|
|  | Best | Worst | Best / (Best + Worst) | Best | Worst | Best / (Best + Worst) |
| Film A | 0.0000 | 1.0000 | 0.0000 | 0.6377 | 1.0000 | 0.3894 |
| Film B | 0.0833 | 0.0000 | 1.0000 | 0.5173 | 0.4830 | 0.5171 |
| Film C | 0.0000 | 0.1250 | 0.0000 | 0.3740 | 0.5527 | 0.4036 |
| Film D | 0.1176 | 0.0000 | 1.0000 | 0.5222 | 0.3993 | 0.5667 |
| Film E | 0.0000 | 0.0588 | 0.0000 | 0.4215 | 0.5900 | 0.4167 |
| Film F | 0.1333 | 0.0000 | 1.0000 | 0.6546 | 0.3867 | 0.6286 |
| Film G | 0.2500 | 0.0000 | 1.0000 | 0.6865 | 0.3747 | 0.6469 |
| Film H | 0.0667 | 0.0323 | 0.6739 | 0.4835 | 0.4444 | 0.5211 |
| Film I | 0.0000 | 0.0000 | UNDEF | 0.9195 | 0.7182 | 0.5615 |
| Film J | 1.0000 | 0.0000 | 1.0000 | 1.0000 | 0.6377 | 0.6106 |

The Jaccard similarity measure (6) is clearly not capable when used as a method of ranking by itself, as all IAA Fuzzy Numbers that do not overlap with the ideal best or worst are not adequately assessed; all evaluated with the lowest equal rank of 0. The issue is further emphasized when an IAA Fuzzy Number is overlapping with neither ideal, such as Film I, which results in a 'division by zero' error as the total distance to both ideals is 0. As it has the ability to evaluate all IAA Fuzzy Numbers, the attribute similarity measure (7) is technically capable of functioning as a method of ranking. However, (7) still does not meet the standard requirements of ranking IAA Fuzzy Numbers that are clearly an ideal best or ideal worst as the highest and lowest ranking respectively, this is again due to its emphasis on pattern matching. Film G is incorrectly ranked as the highest as it is very close to the actual ideal best, Film J, and it does not share structure with the ideal worst, Film A. Film A and Film J are identical in shape, they are just placed on opposite boundaries of the x-axis. As Film J is the ideal best but shares structure with the ideal worst, it consequently results in a ranking lower than the highest when applying (7), which violates the requirements.

IV. PROPOSED METHODS

In this section, we propose and outline a new similarity measure and two ranking methods for IAA Fuzzy Numbers.

*A. Ranking Method: Universal*

This ranking method primarily utilizes the centroid (center of mass) and in some cases the perimeter attributes for IAA Fuzzy Numbers as proposed in [9]. We refer to this ranking method as universal as given any two IAA Fuzzy Numbers it computes if one is greater or if they equal regardless of context. The exclusion of context is this ranking method's main advantage and is the reason for its proposal. Gunn et al. [9] propose a refined centroid equation that equally weighs certainty and uncertainty in the case of IAA Fuzzy Numbers, given as (8, 9, 10), where *n* is equal to the length of $MC_A$.

$$C_x(MC_A) = \frac{\sum_{i=0}^{n}(R_{i_h} \times R_{i_l}) + (R_{i_h} \times R_{i_l})}{\sum_{i=0}^{n} 2(R_{i_h})} \quad (8)$$

$$C_y(MC_A) = \frac{\sum_{i=0}^{n} \frac{R_{i_h}}{2}}{non\_zero(R_{i_h})} \quad (9)$$

$$non\_zero(x) = \begin{cases} 1 & x > 0, \\ 0 & else. \end{cases} \quad (10)$$

The outline for the universal ranking method given two IAA Fuzzy Numbers $\tilde{A}$ and $\tilde{B}$ is as follows:

1. The IAA Fuzzy Number with the greater centroid-x outranks the other. If the centroid-x values are equal, proceed to step 2.
2. The IAA Fuzzy Number with the lower perimeter outranks the other, and this is because lower values of perimeter imply the fuzzy number is more certain than the other and is thus relatively risk-averse and more reflective of the centroid-x. If the perimeters are equal, proceed to step 3.
3. The IAA Fuzzy Number with the greater centroid-y outranks the other. If the centroid-y values are equal, proceed to step 4.
4. This ranking method assumes both IAA Fuzzy Numbers $\tilde{A}$ and $\tilde{B}$ are equal in rank.

Note that a similar ranking method for Generalized Fuzzy Numbers is explored in [13], applying different attribute outlines that were fitting to that data type.

*B. Ranking Method: Ideal Similarity*

This ranking method combines both similarity measures outlined in Section II, thus proposing a new measure, and through its application includes a rudimentary means of ranking. As explored in Section III, both similarity measures (6) and (7) have shortcomings when applied to ranking. Intriguingly, both the similarity measures (6) and (7) address the limitation of the other. The Jaccard similarity coefficient measure (6) provides a higher weighting to IAA Fuzzy

Numbers that overlap with an ideal, and the attribute similarity measure (7) provides a means of computing similarity when no overlap occurs. In an effort to capture the benefits of each similarity measure, the modification we propose is combining the Jaccard similarity coefficient measure (6) with the attribute similarity measure (7).

The combined similarity measure is given by (11), for the purposes of this study, it is a basic ensemble method in which the outputs of each previous measures are averaged.

$$S(\tilde{A}, \tilde{B}) = \left(\frac{S_J^{CS}(\tilde{A},\tilde{B}) + S_{AC}(\tilde{A},\tilde{B})}{2}\right) \quad (11)$$

A ranking method for IAA Fuzzy Numbers with the inclusion of both the ideal best and ideal worst is now given by (12), in which higher outputs imply a higher ranking. Note that the ideal best IAA Fuzzy Number in the numerator of (12) can be exchanged for the ideal worst, in doing so lower outputs will now imply a higher ranking.

$$Rs(\tilde{A}, \tilde{I}_b, \tilde{I}_w) = \frac{S(\tilde{A},\tilde{I}_b)}{S(\tilde{A},\tilde{I}_b) + S(\tilde{A},\tilde{I}_w)} \quad (12)$$

### C. Illustrated Ranking Improvements

Returning to the synthetic example previously explored in Section III and originally provided by [9], we now apply the proposed ranking methods as given by Table. V.

TABLE V. FINAL RANKING - FILM REVIEW DATASET

| Method | $Rs(\tilde{F}, \tilde{I}_b, \tilde{I}_w)$ -Eq. (12) | $Rs(\tilde{F}, \tilde{I}_b, \tilde{I}_w)$ -Eq. (12) | Universal Ranking | Preprocess Averaging |
|---|---|---|---|---|
| FORMAT | Original Output | Ranking | | |
| Film A | 0.2418 | 10 | 10 | 10 |
| Film B | 0.5543 | 6 | 7 | 6 |
| Film C | 0.3556 | 9 | 9 | 9 |
| Film D | 0.6157 | 4 | 5 | 5 |
| Film E | 0.3938 | 8 | 8 | 8 |
| Film F | 0.6708 | 3 | 3 | 3 |
| Film G | 0.7142 | 2 | 2 | 2 |
| Film H | 0.5358 | 7 | 6 | 7 |
| Film I | 0.5615 | 5 | 4 | 4 |
| Film J | 0.7582 | 1 | 1 | 1 |

Both proposed ranking methods only differ with the more traditional preprocessed averaging method by a factor of one switch within the ordering each, both of which only by one rank. The attribute similarity measure (7) was included to emphasize both the necessity of including the ideal worst as well as the ideal best, and a push towards IAA Fuzzy Numbers that overlap [7] with them, of which a clear improvement has been observed. The combined similarity ranking method (12) switches the rank of Film D and Film I, this is due to Film I's reflective IAA Fuzzy Number not overlapping with neither the ideal best nor ideal worst, yet still having a very similar shape to both regardless (single line), thus causing it to be more centered than within the other two methods. The universal ranking method switches Film B and Film H, and this logically follows as their centroid-x scores (of which the universal ranking method primarily utilizes) were 5.9375 and 6.0364 respectively. If Film H's corresponding IAA Fuzzy Number was slightly more skewed to the left (refer to Fig. 2), it would be considered lower than Film B. Hypothetically even if both films were equal, by the default separation method outlined, Film B having the lower perimeter would be considered higher ranking.

Considering that Film B and Film H have very similar values for centroid-x, if a scenario were to call for it, it would be feasible to consider putting a threshold range that the fuzzy numbers may be within (in contrast to their values just being equal) in order to utilize risk vs opportunistic models. The outcome of the proposed ranking methods appears desirable. Being exactly the same as the more traditional averaging approach would prove the procedure of using interval-valued data and IAA Fuzzy Numbers to likely be unbeneficial, however, when given uncertain values these two proposed procedures output very similar results with only slight variations; thus they are very well within the bounds of expectation whilst potentially providing novel inferences.

Further modification to the combined similarity ranking measure (12) in order to achieve the same results as the traditional averaging method is certainly possible, such as by changing the weights of features or inner measures; though doing so at this stage of research may be considered arbitrary and intriguing inferences regarding the uncertainty within the data may be lost.

## V. CASE STUDY - CYBER-SECURITY EVALUATION

The data utilized for this example application of a real-world scenario was primarily gathered from [14], with recent applications also utilizing it in [15]. Due to confidentiality concerns, as well as the size of the dataset, we are unable to provide a full table of the original data collection. However, it identical in structure to the synthetic dataset previously outlined in Section III, with the significant difference being that it includes multiple criteria, that is it is a multi-dimensional variant of a very similar scenario. For the synthetic film dataset, only the overall review has been hypothetically gathered. If opinions for other criteria were to also be gathered from critics, such as for example acting, music, visual effects, sound design, comedy, maturity, etc, this would produce a multi-criteria dataset.

The Cyber-security dataset is a survey regarding the quality of technical attacks. Cyber-security experts were asked to provide intervals within the range of 0 to 100, of their opinion of various criteria regarding the quality of 14 technical attacks, as well as their overall opinion regarding them. The 9 criteria used were as follows: attack, complexity, interaction, frequency, availability of tool, inherent difficulty, maturity going unnoticed, and overall. Khorshidi and Aickelin [15] proposes a multicriteria group decision-making (MCGDM) algorithm, which applies the Technique for Order of Preference by Similarity to Ideal Solution (TOPSIS) algorithm [17] (note TOPSIS does not utilize the overall opinion criteria), and compares the results to two other algorithms found within [1] and [18].

TOPSIS is a method of ranking multi-criteria data, and within its outlined steps it requires both a similarity and ranking method for input data types, opening an opportunity for the application of methods proposed within this study. Following [15] the overall opinion has been removed, and each criteria interval set regarding each attack provided by each expert were converted to IAA Fuzzy Numbers and passed into TOPSIS utilizing the attribute similarity measure (7) and the proposed combined similarity measure (11) as a means of contrast; both of which had the proposed universal ranking method applied when required by TOPSIS. Having the overall criterion removed gives the opportunity to compare the proposed universal ranking method with the results of the TOPSIS combination. The IAA Fuzzy Numbers produced from the overall criterion have been used as input for the universal ranking method. For further details regarding the TOPSIS outline refer to [15, 17].

The proposed IAA similarity and ranking methods were successfully combined with the multi-criteria ranking approach of TOPSIS, refer to Fig. 3.

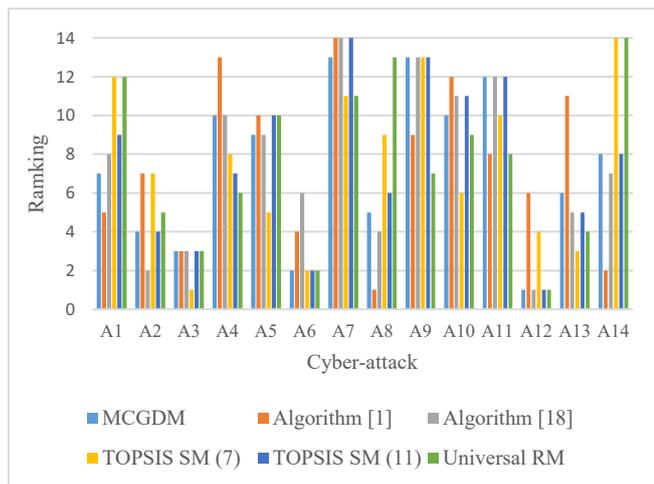

Fig. 3. Output of various ranking measures

IAA Fuzzy Numbers constructed from interval sets aggregated from the Cyber-security dataset [14] being used as input to TOPSIS, along with the proposed universal ranking method do appear reasonable when compared to the other methods explored in [15]. Intriguingly, all of the algorithms compared do vary in their results, and as such, there is undoubtedly room for further investigating the effectiveness of IAA combined with TOPSIS. Moreover, the primary focus of [15], the recent MCDGM algorithm shares strikingly similar results with the proposed combined similarity measure (11), which shows promise for its potential effectiveness.

The Cyber-security data utilized within this section was recorded in an unsupervised fashion, thus how these algorithms should be skewed has not been fully determined as of the writing of this paper. However, as is the primary goal of including this case study and of this paper itself, the ability to apply the proposed similarity and ranking methods and acquire fitting results demonstrates the potential usefulness in future research.

## VI. Conclusion

This paper has proposed both a new similarity measure and two ranking methods for IAA Fuzzy Numbers, which were the primary novel findings of the research involved. Acquiring interval scores in contrast to real numbers and applying the various techniques explored throughout was indeed shown to be an effective alternative that has the ability to make reasonable inferences, whilst maintaining the uncertainty of the input data throughout. The output of the proposed methods was in multiple instances very similar to other non-fuzzy methods, which emphasizes the conjecture that IAA Fuzzy Numbers may be comparably effective to other common standards of data analysis even when provided uncertain or vague input.

For future work, there are many clear modifications that may be implemented for the methods proposed by this paper. Much of this study is a proof of concept and as such the methods used throughout are rudimentary. For example, the Jaccard similarity coefficient measure is currently only applied in a very discrete manner, and therefore an update to that measure that accounts for more of the input IAA Fuzzy Numbers would yield more effective results. The proposed combined similarity measure is currently a basic ensemble method, and thus additional complexity regarding the weights of its inner measures and how said inner measures collaborate would likely improve its potential for application. The abstraction of this study's findings into Type-2 Fuzzy Numbers is also a potentially useful large-scale modification, as much modern research continues to apply them.

The IAA method was developed to capture and maintain the uncertainty of data sources throughout its application, and with this in mind, its use as a data preprocessor could be applied to many Machine Learning models. A difficulty with applying IAA Fuzzy Numbers is that they do not have well-defined methods of comparison, relative to alternative data types. This study addresses a portion of these necessary methods, allowing for further analysis of the application of IAA Fuzzy Numbers as an alternative input to many statistical models. It would be intriguing to analyze the change in effectiveness and potentially unique output of popular Machine Learning algorithms with the additional application of the IAA method for capturing uncertainty.